\documentclass[pdflatex,sn-mathphys,Numbered]{sn-jnl}

\usepackage{graphicx}%
\usepackage{multirow}%
\usepackage{amsmath,amssymb,amsfonts}%
\usepackage{amsthm}%
\usepackage{mathrsfs}%
\usepackage[title]{appendix}%
\usepackage{xcolor}%
\usepackage{textcomp}%
\usepackage{manyfoot}%
\usepackage{booktabs}%
\usepackage{algorithm}%
\usepackage{algorithmicx}%
\usepackage{algpseudocode}%
\usepackage{listings}%

\usepackage{hyperref}
\usepackage{comment}

\usepackage{amsmath}
\usepackage{amsfonts}

\DeclareMathOperator*{\argmin}{arg\,min}

\usepackage{eqparbox}

\newcommand{\rev}[1]{\textcolor{black}{#1}}

\usepackage{xcolor, soul}

\theoremstyle{thmstyleone}%
\theoremstyle{thmstyletwo}%

\theoremstyle{thmstylethree}%

\begin{document}

\title[Article Title]{Beyond the Manual Touch: Situational-aware Force Control for Increased Safety in Robot-assisted Skullbase Surgery}




\author*[1]{\fnm{Hisashi} \sur{Ishida}}

\author[1,2]{\fnm{Deepa} \sur{Galaiya}} 

\author[2]{\fnm{Nimesh} \sur{Nagururu}} 

\author[1,2]{\fnm{Francis} \sur{Creighton}}

\author[1]{\fnm{Peter} \sur{Kazanzides}} 

\author[1,2]{\fnm{Russell} \sur{Taylor}} 

\author*[1]{\fnm{Manish} \sur{Sahu}}\email{manish.sahu@jhu.edu}

\affil[1]{\orgdiv{LCSR}, \orgname{Johns Hopkins University}, \orgaddress{\country{USA}}}

\affil[2]{\orgdiv{Department of Otolaryngology}, \orgname{Johns Hopkins University}, \orgaddress{\country{USA}}}


\abstract{
\textbf{\\* Purpose}:
Skullbase surgery demands exceptional precision when removing bone in the lateral skull base. Robotic assistance can alleviate the effect of human sensory-motor limitations. However, the stiffness and inertia of the robot can significantly impact the surgeon's perception and control of the tool-to-tissue interaction forces.
\textbf{\\* Methods}: 
We present a situational-aware, force control technique aimed at regulating interaction forces during robot-assisted skullbase drilling.
\rev{The contextual interaction information derived from the digital twin environment is used to enhance sensory perception and suppress undesired high forces.}
\textbf{\\* Results}: 
To validate our approach, we conducted initial feasibility experiments involving \textcolor{black}{a medical and two engineering students}. The experiment focused on further drilling around critical structures following cortical mastoidectomy. The experiment results demonstrate that robotic assistance coupled with our proposed control scheme effectively limited undesired interaction forces when compared to robotic assistance without the proposed force control.
\textbf{\\* Conclusions}: 
The proposed force control techniques show promise in significantly reducing undesired interaction forces during robot-assisted skullbase surgery. These findings contribute to the ongoing efforts to enhance surgical precision and safety in complex procedures involving the lateral skull base.
}

\keywords{Tool-tissue interaction, Cooperative control, Adaptive force control, Force feedback, Skullbase surgery, Surgical robotics, Digital twin}



\maketitle

\section{Introduction}\label{sec:intro}

Skull base surgery is technically challenging and requires precision in the execution of intricate tasks within the constrained and delicate lateral skull base environment.
To access this region, surgeons must skillfully drill through varying densities of bone to reveal critical anatomical structures, often hidden by operable tissue at sub-millimeter distances~\cite{meybodi2023comprehensive,jain2019anatomical}. Microsurgery is constrained by the limits of human visual, tactile, and motor control thresholds~\cite{kratchman2016force}, making the tissue manipulation a challenging endeavor. Consequently, surgeons undergo a rigorous minimum of 7 years~\cite{gantz2018evolution}  of postgraduate training to master the complexities of this anatomical region.

One significant challenge lies in the subtle nature of tool-to-tissue interaction forces, which can fall at or below human tactile perception levels\cite{trejos2010force}. The introduction of robotic assistance holds promise in enhancing surgical performance by enhancing tool tip precision and reducing the impact of hand tremors during delicate surgical tasks. However, while robots can provide steady-hand manipulation, the sensory feedback traditionally derived from manual tool interactions is altered, potentially leading to unintended contact between the instrument and tissue, with the potential for serious injury.
The accurate sensing and intelligent control of tool-tissue interaction forces are pivotal in enhancing surgical safety and reducing the incidence of complications~\cite{simaan2015intelligent,attanasio2021autonomy}. The success of robotic assistance in skull base procedures hinges on the precision of the sensing and control mechanisms that adapt to the surgical context and ensure that tool-tissue interactions remain within safe thresholds.


In this work, we address these challenges through the development of a situation-aware, adaptive control method tailored for controlled tissue ablation in skull base procedures. \rev{Our control method leverages contextual information from the dynamically changing surgical environment to regulate interaction forces concerning desired forces related to distinct anatomical structures. Given the tendency of medical trainees to apply more force than attending surgeons, contributing  to technical errors~\cite{golahmadi2021tool}, we hypothesize that our proposed situational-aware adaptive force control strategy can reduce undesired contact forces during collaborative drilling in skull base procedures.}


\rev{To validate our hypothesis, we conducted experiments with two attending surgeons, performing an initial mastoidectomy around various anatomical structures. The resulting drilling values were statistically analyzed to determine expected forces during these procedures. Subsequently, safety threshold values for the force control scheme were established based on these statistical insights.
To demonstrate the effectiveness of our proposed control scheme, we conducted a cadaveric temporal bone experiment involving three inexperienced users. The experiment included drilling around critical structures following cortical mastoidectomy. A comparative analysis of our proposed method against cooperative robotic assistance without force control employed a comprehensive set of force metrics~\cite{golahmadi2021tool}, including average forces, maximum forces, and the time spent over threshold force. The experimental results consistently demonstrate that the active robot control schemes maintain applied forces below the desired safe threshold during robot-assisted skull base surgery.}

\rev{The contributions of our work include: 1) the extension of the Digital Twin (DT) environment to incorporate contact interactions through force measurements, 2) the development of an approach for accurate estimation of the currently operated anatomical structure based on contextual information in DT environment, and 3) the development of an adaptive force control technique for controlled tissue ablation in skull base procedures.}

\section{Related Work}\label{sec:related_work}



\rev{Traditional methods in robot-environment interaction are commonly classified into two main categories: indirect force control (without explicit force feedback) and direct force control \cite{haidegger2009force,suomalainen2022survey}.
Indirect force control involves performing manipulation tasks by leveraging the interplay between motion and interaction forces. However, a potential drawback is the risk of losing contact with the environment due to the absence of explicit force feedback \cite{minniti2021model}.
In contrast, direct force control facilitates manipulation by explicitly tracking a desired force, making it particularly effective for surgical tasks where specific force profiles are essential.
However, the efficacy of direct force control is contingent upon prior knowledge of the surgical task and a situational understanding of the dynamic, time-varying surgical environment.}

In the pursuit of enabling robotic assistance in surgery, prior work~\cite{trejos2010force} in the literature has predominantly focused on advancing hardware and sensor technologies to facilitate steady-hand manipulation and reduce the impact of hand tremors. However, comparatively less attention has been directed toward the development of control and sensing algorithms.\cite{golahmadi2021tool}
\rev{In the realm of orthopedic surgery, studies propose force control approaches to enhance positional accuracy of a hand-held robot \cite{yen2022contact}. Also, force sensing methods have been developed for tasks such as screw-path-drilling \cite{ying2014state} and cooperatively assisted surgical needle insertion\cite{li2022admittance}. Additionally, a force-control-based approach is developed for regulating penetration force during cell injection in zebrafish embryos \cite{xie2010force}.}

Within the context of ENT surgery, Rothbaum et al.~\cite{rothbaum2002robot, rothbaum2003task} evaluated the potential of robotic assistance in stapedotomy and showed that it can significantly limit the maximum force applied to the stapes footplate.
Ebrahimi et al.\cite{ebrahimi2019adaptive} proposed control methods for eye surgery to limit sclera forces applied to the tool shaft at the trocar, further showcasing the adaptability of robotic systems in surgical contexts.

Sang et. al.~\cite{sang2017new} integrated a force sensing drill to the end effector of a da Vinci robot and showed its utility for robotically assisted otologic surgery. Additionally, systems like MMS~\cite{maier2010mms}, utilizing a control console with two joysticks, and RobOtol~\cite{miroir2010robotol}, which employs a Phantom Omni joystick for force feedback, have been explored. However, these approaches largely follow the teleoperation paradigm, introducing challenges such as removing surgeons from direct patient access and necessitating substantial alterations to conventional surgical procedures.
In contrast, cooperative systems have emerged as a cost-effective and minimally disruptive alternative. These systems offer intuitive control over teleoperation, requiring less training for surgeons to become proficient. Furthermore, they enable surgeons to remain at the patient's bedside, enhancing the overall surgical experience.

In our prior work~\cite{chen2023force}, we have developed a force-sensing surgical drill for a hand-over-hand cooperatively controlled surgical robot that can measure tool-to-tissue forces with millinewton accuracy in real-time.
\rev{This paper extends our previous efforts, focusing on the development of adaptive force control schemes to actively constrain interaction forces within the surgical environment.}
\rev{This involves seamless integration of patient anatomy, the cooperative robot, and the interaction forces into a real-time DT environment as well as utilization of the contextual information in DT to dynamically adjust the control parameters.}
In a recent study~\cite{ishida2024haptic}, we introduced safety-driven virtual fixtures intended to guide drill motion away from critical structures. However, these fixtures did not account for the impact of tool-tissue interaction forces. In contrast, this paper places a distinct focus on the real-time interaction forces, as well as the development of situation-aware active force control aimed at constraining interactions with the surgical environment.


\section{Methods}\label{sec:methods}

In the following section, we describe the proposed situational-aware force control approach. The main motivation for this method is to balance high transparency during non-contact periods and ensure safer tool-tissue interaction during contact. The proposed method combines the real-time environment information (tool-tissue interaction force) and surgical context (operating anatomical structure and safety margin for the structure) from the virtual simulator to enhance surgical safety.
\rev{First, the importance of combining the environment information with surgical context is described in Sec. \ref{sec:Digital Twin}.}
The cooperative robotic system is described in Sec. \ref{sec:REMS}, and the main control scheme is described in Sec. \ref{sec:Control}. 

\begin{figure}[th]%
\centering
\includegraphics[width=1.0\textwidth]{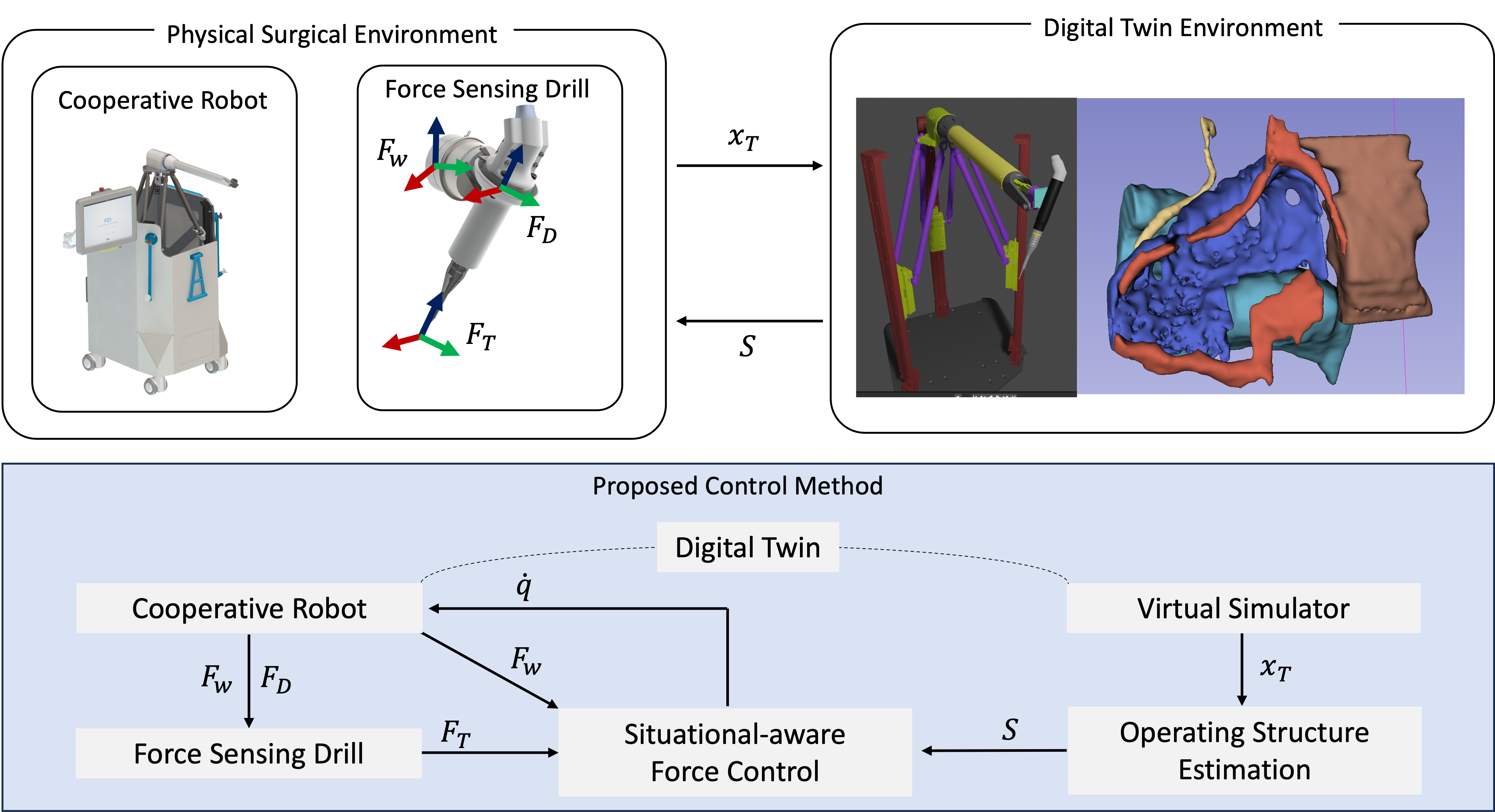}
\caption{System overview showing force sensing drill, surgical environment, and digital twin. $S$ denotes the operating anatomical structure. $x_T$ is the drill tip pose and $\Dot{q}$ is a joint velocity. $F_T$, $F_W$, $F_D$ denote tool-tissue interaction force at the drill tip, and forces measured from the wrist force sensor and the drill sensor, respectively.}
\label{fig:SystemOverview}
\end{figure}

\subsection{\rev{Digital Twin with Tool-tissue interactions}}
\label{sec:Digital Twin}

\rev{The efficacy of any shared control functionality within the surgical context necessitates a comprehensive understanding of the environment and the ongoing task. Leveraging a DT approach holds the promise of deriving semantic knowledge about the surgical environment and the system's state. Consequently, we integrated a DT environment tailored for skull base procedures \cite{shu2023twin}. The development of the DT framework mandates precise modeling of individual system components to closely emulate a real surgical environment. Intraoperatively, these DT models undergo registration with the actual patient, enabling the tracking of surgical instrument positions relative to the patient. However, due to the inherent inaccuracies of current tracking systems, relying solely on spatial information proves inadequate for precisely identifying contacts with the anatomy.
In this work, we enhance the DT system by integrating force sensing measurements~\cite{chen2023force}, crucial for establishing a real-time and precise understanding of tool-tissue interactions. This extension ensures a more nuanced and accurate representation of the surgical scenario, acknowledging the dynamic interplay of spatial information and interaction forces.}

\subsection{Cooperative Robot}
\label{sec:REMS}
The Robotic ENT (Ear, Nose, and Throat) Microsurgery System (REMS) is a cooperatively-controlled robot specialized for use within otolaryngology–head and neck surgery~\cite{olds2014preliminary,olds2015phd}. In this study, we use a pre-clinical version developed by Galen Robotics (Baltimore, MD). One of the primary advantages of REMS is that it offers a significant benefit in instrument stability in head and neck surgery. The cooperative mode of the robot uses the admittance control~\cite{keemink2018admittance} law:

\begin{equation}
\argmin_{\Delta q} \left( |GF_{H} - J\Delta q|\right)
\end{equation}

$G\in \mathbb{R}^{6 \times 6}$ denotes a diagonal matrix that represents the admittance gains, $J \in \mathbb{R}^{6 \times m}$ is a Jacobian and $\Delta q \in \mathbb{R}^m$ is a joint velocity vector. The incremental robot motion, $\Delta x \in \mathbb{R}^6$, is expressed as ($\Delta x = J \Delta q$). 

To deliver a sensation of touch and avoid damaging the anatomy during contact without disturbing the surgeons' operation, we introduce a gain adjustment term, $ \sigma \in \mathbb{R}$. Consecutively, the robot is controlled by the following admittance law. $G' = \sigma I G$ is an adjusted admittance gain, where $I \in \mathbb{R}^{6 \times 6}$ is an identity matrix.

\begin{equation}
\argmin_{\Delta q} \left( |G'F_{H} - J\Delta q|\right)
\end{equation}

We increase $\sigma = \sigma_{high}$ ($\sigma_{high} > 1.0$) so that the robot will respond more promptly to the human input when there is no contact with the anatomy ($|F_{contact}| < C$), where $C$ is a contact threshold. Meanwhile, during the contact, $\sigma$ is set as $\sigma_{contact}$ ($\sigma_{contact} < 1.0$) to ensure precise and delicate motion.

\subsection{Control schemes}
\label{sec:Control}

\rev{This section describes the adaptive gain control scheme proposed in this paper. 
The back end module involves the situation-aware Digital Twin environment to identify the operating structure. The front end consists of adaptive force control to regulate the interaction between the surgical and the patient anatomy.}

\subsubsection{Operating structure estimation}
\label{sec:Operating_Structure_Estimation}
Given the delicate nature of each anatomical structure, precise regulation of interaction forces tailored to the specific anatomy is imperative.
The real-time computation of the minimum distance between the drill tip and its nearest anatomical structure can be acquired using signed distance fields~\cite{ishida2023improving}.
Given $n\in \mathbb{Z}$ anatomical structures of the patient model, one can determine the distance $d_n \in \mathbb{R}$ between the drill tip and the $n^{th}$ anatomical structure. 

\rev{The operating structure, denoted as $\mathbb{S}$, is identified based on the proximity to the drill tip and real-time interaction force measurements, which are continuously tracked by the digital twin paradigm described in Section \ref{sec:Digital Twin}. If the closest distance to an anatomical structure, $d_{min} \leq \forall d_n$, is less than a predefined threshold, $\gamma_{n}$, and the force value surpass contact threshold, we recognize the closest structure as the operating structure.}


\begin{figure}[h]%
\centering
\includegraphics[width=0.5\textwidth]{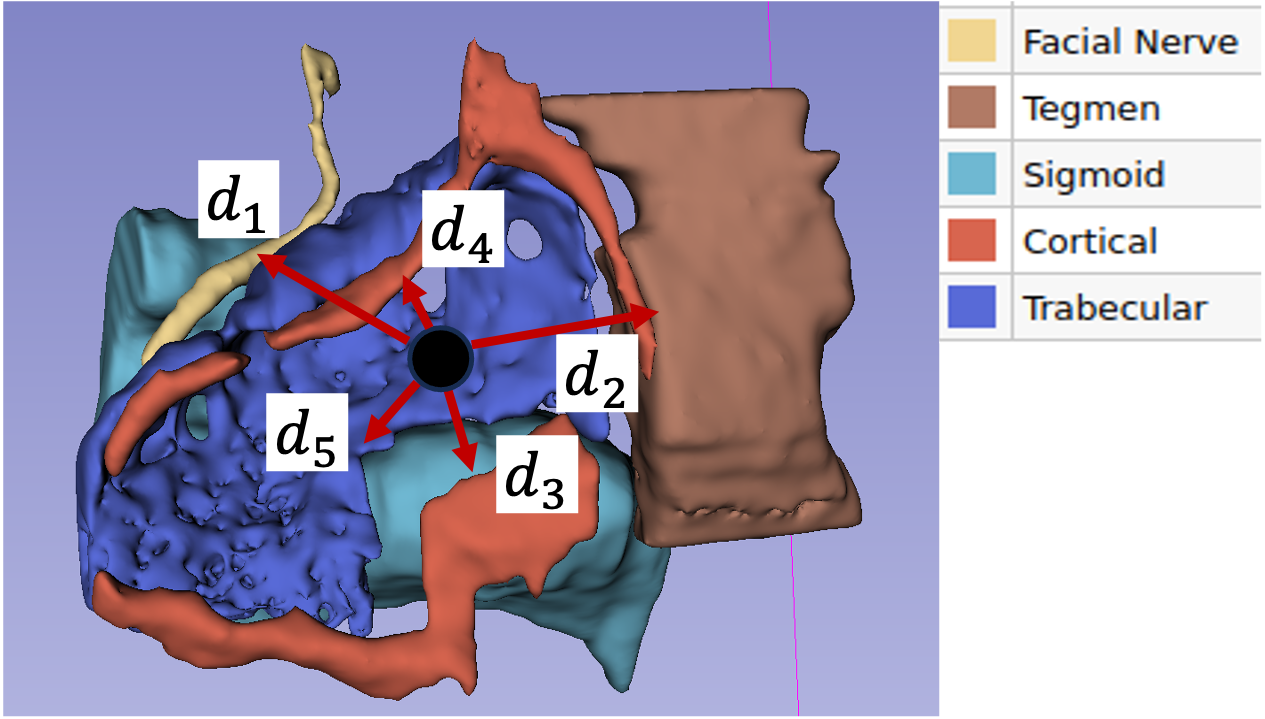}
\caption{The closest distances between the drill tip and the anatomical structures, $d_n$, are calculated in real-time. ($d_1$: Facial Nerve, $d_2$: Tegmen, $d_3$: Sigmoid, $d_4$: Cortical bone, $d_5$: Trabecular bone). \label{fig:StepRecognition}}
\end{figure}

For each operating structure, a safety threshold denoted as $\lambda_n \in \mathbb{R}$ is established as an acceptable interaction force. This allows the proposed force control to incorporate both real-time interaction forces with the tissue and surgical context information.

\subsubsection{Situational-aware force control}
To minimize the undesired force applied to the anatomy, we propose a situational-aware force control method that adaptively adjusts the gain using both the tool-tissue interaction force and the operating structure. \rev{The control gains remain consistently high ($\sigma_{high}$) to increase the transparency of the robot and adjusted when making contact with the anatomy to enhance the sensation of touch. 
The gain adjustments are introduced during anatomical ablation. The adaptive gains remain unchanged ($\sigma_{contact}$) until force values exceed safe force ranges ($C < |F_{contact}| < U$). Once the force values surpass undesired ranges, the control method adjusts gains gradually, making the robot increasingly stiff (Eq. \ref{eq:sigma}). This adjustment is proportional to the overall time spent beyond the desired force values. }
This gain adjustment term, $\sigma(t)$ can be expressed as 
\begin{equation}
\label{eq:sigma}
    \sigma(t) = 
    \begin{cases}
    \sigma(0)\exp\left(-\eta \Delta F(t-t_0)\right) + \sigma_{low} & \text{if $|F_{contact}| \rev{\geq} U$} \\
    \rev{\sigma_{contact}} & \text{if $U > |F_{contact}| \rev{\geq} C$}\\
    \sigma_{high} & \text{if $|F_{contact}| < C$}
    \end{cases}
\end{equation}
\rev{where $U$ is an anatomy-specific threshold for undesired force derived in Sec. \ref{sec:Operating_Structure_Estimation} $(U = \lambda_n)$, and C is the contact threshold.} $\sigma(0) = \sigma_{contact} - \sigma_{low}$ and $\Delta F = F_{contact} - U$. $t_0$ is the time when contact force, $F_{contact}$, surpasses, \rev{$U$.} $\eta \in \mathbb{R}$ represents a decaying constant. $\sigma_{low} \in \mathbb{R}$ ($\sigma_{low} < \sigma_{contact}$) is a lower bound gain adjustment term that mitigates further anatomical contact. 
\rev{This gain adjustment term is used in admittance control explained in Algorithm \ref{alg:adaptive control}, which hinders further damage to the tissue while maintaining the control to follow the user input.}
\rev{
\begin{equation}
\label{eq:adjusted_eq}
\argmin_{\Delta q} \left( |\sigma(t)IGF_{H} - J\Delta q|\right)
\end{equation}
}

\begin{algorithm}[h]
\caption{Situational-aware force control}\label{alg:adaptive control}
 \hspace*{\algorithmicindent} \rev{\textbf{Input:} State of the environment: \\User input $F_{H}$, Tool-tissue interaction force $F_{T}$, drill tip pose w.r.t anatomy $\Vec{x}_{T}$} \\
 \hspace*{\algorithmicindent} \rev{\textbf{Output:} Robot motion $\Delta q$}
\begin{algorithmic}

\If{$|F_T|> C$}\Comment{Once there is a contact}
    \State $S \gets \mathrm{getOperatingStructure}(\Vec{x}_T)$\Comment{Get operating structure}
    \State $\rev{U}  \gets \mathrm{getInfo}(S)$  \Comment{Get undesired force threshold}
    \State $\sigma = \sigma_{contact}$ 
\While{ $|F_T| > \rev{U}$} \Comment{Once undesired force applied}
    \State Calculate $\sigma(t)$ based on equation \ref{eq:sigma}
\EndWhile
\Else \Comment{During freehand motion}
    \State $\sigma = \sigma_{high}$ 
\EndIf

\rev{Calculate $\Delta q$ with proposed admittance control (equation \ref{eq:adjusted_eq})}
\State \textbf{return} $\Delta q$
\end{algorithmic}
\end{algorithm}

The overall algorithm that summarizes the proposed method can be found in Algorithm \ref{alg:adaptive control}. Since the operating structure and its acceptable interaction force, $\lambda_n$, are estimated from Section \ref{sec:Operating_Structure_Estimation}, gain adjustment term, $\sigma(t)$, changes its value in real time incorporating the spatial information and the surgical context. \rev{Consequently, realizing a situational-aware adaptive force control strategy which can
reduce undesired interaction force during tissue ablation.}

\section{Experiments}\label{sec:experiments}


We conducted a feasibility study to demonstrate our method's application and to gather initial feedback for a larger user study in the future.

\subsection{Experiment setup}
\begin{figure}[t]%
\centering
\includegraphics[width=0.65\textwidth]{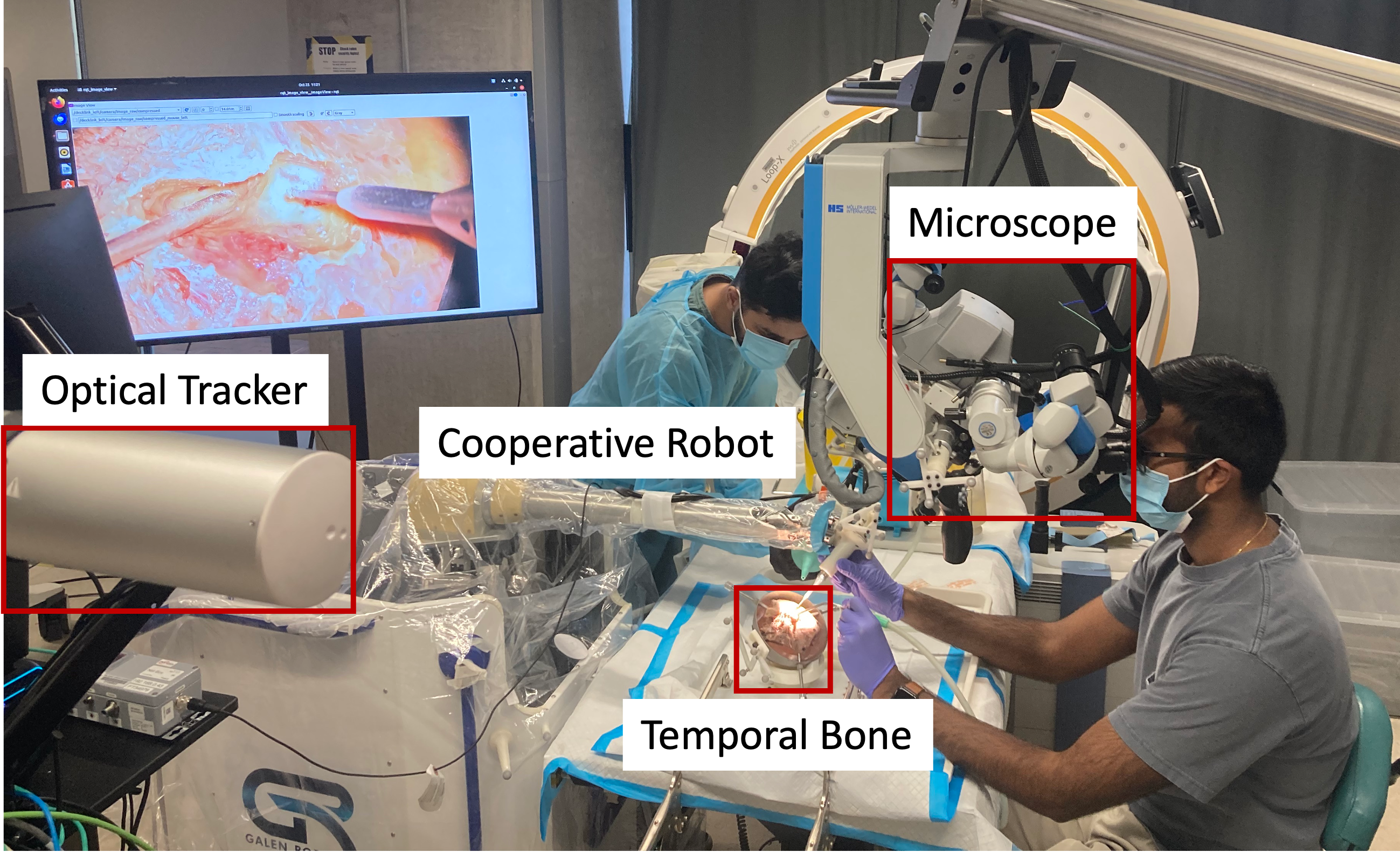}
\caption{Experimental setup. Medical student uses the surgical drill attached to the cooperative robot under microscopic view, while an adjacent optical tracker monitors both the drill and the anatomy.}\label{fig:Experimental_Setup}
\end{figure}

In preparation for our experiment, we attached registration pins to the cadaveric temporal bone. The bone was then scanned using high-resolution CBCT (Brainlab LoopX, $0.26\,mm^3$ voxel size); scans were then taken for each temporal bone. Using 3D Slicer, we annotated critical structures and pin locations. These annotations were instrumental in crafting the patient's anatomical model for our virtual simulator.
Prior to the experiment, the temporal bone was securely mounted in a designated holder, which was then anchored to a surgical table to eliminate unintentional movement. To track movements during the experiment, markers were placed on the surgical drill and the bone holder.  Following this, we used a 2 $mm$ drill tip for pivot calibration, achieving an RMSE value of 0.1 $mm$.
Finally, a point-set registration method was performed to register the temporal bone to the virtual simulation. These points were carefully sampled using the surgical drill, leading to an RMSE of less than 0.5 $mm$.
During the experiment, drilling procedures was conducted using a surgical microscope (Haag-Streit, Köniz, Switzerland) equipped with stereo vision (Fig. \ref{fig:Experimental_Setup}).
\rev{Force sensors are sampled in 200 $Hz$, collaborative robot and virtual simulation are running in 500 $Hz$ and 1000 $Hz$ respectively.}

\subsection{Experiment design}



The experiment was carefully designed to specifically address the heightened relevance of force control feedback in the later stages of skull base procedures, where excessive force can have critical implications.
The initial mastoidectomy was conducted by two attending surgeons to derive expected forces during drilling around various anatomical structures, including cortical bone, trabecular bone, and critical structures such as the tegmen, sigmoid, and facial nerve. The statistical values obtained from drilling were then used to establish safety threshold values for the force control scheme, with the statistical mean chosen as the desired force.
To ensure a desired safety margin for each individual anatomical structure and afford the robot sufficient time to prevent interaction forces from reaching safety limits, the control methods were activated 0.2 N before the safety margin limit. This activation threshold served as the trigger for initiating the adaptive drill force control, as outlined in Table \ref{table:anatomy}. The gain settings for interaction force concerning different bone types were determined through a preliminary experiment, as detailed in Table \ref{table:param}.
Subsequently, the task of further skeletonizing the anatomical structures was entrusted to \rev{one medical student and two engineering students with no prior experience in surgical drilling}.

\begin{table}[t]
\caption{Proposed control related and anatomical structure related parameters}\label{table:anatomy}\label{table:param}%
\begin{tabular}{p{0.2\textwidth}p{0.2\textwidth}p{0.2\textwidth}p{0.08\textwidth}}
\toprule
$\sigma_{high}$ & $\sigma_{contact}$ & $\sigma_{low}$ & $\eta$\\
\midrule
1.7    & 0.7  & 0.3  & 1.0 \\
\botrule
\end{tabular}
\begin{tabular}{p{0.2\textwidth}p{0.2\textwidth}p{0.2\textwidth}p{0.08\textwidth}}
\toprule
index & name  & $\gamma_n$ [mm] & $\lambda_n$ [N]\\
\midrule
1    & Facial Nerve   & 1.5 & 0.8  \\
2    & Tegemen   & 1.5  & 0.8 \\
3    & Sigmoid   & 1.5  & 0.8 \\
4    & Cortical   & 0.0  & 1.3 \\
5    & Trabecular   & 0.0  & 1.3 \\
\botrule
\end{tabular}
\end{table}

\begin{figure}[b]%
\centering
\includegraphics[width=.9\textwidth]{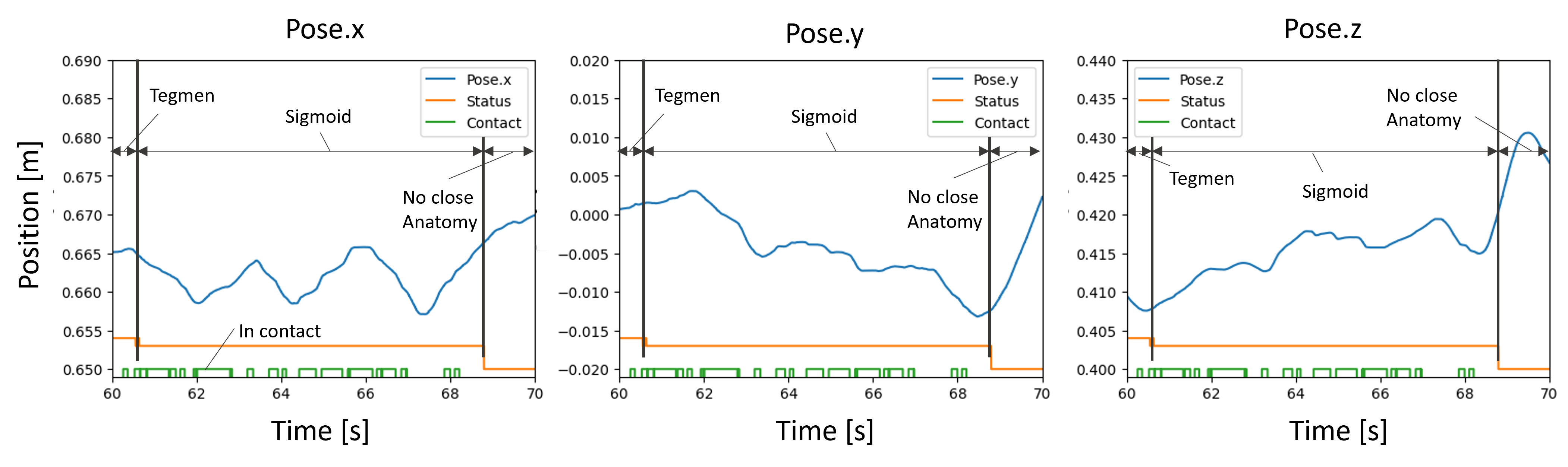}
\caption{\rev
{Positional data (x, y, z) of the drill from a participant's procedure, with a green line indicating drill contact with anatomy (High signifies contact) and labeled operating structures.}}\label{fig:JitterResult}
\end{figure}

\subsection{Result and discussion}

\begin{figure}[t]%
\centering
\includegraphics[width=1.0\textwidth]{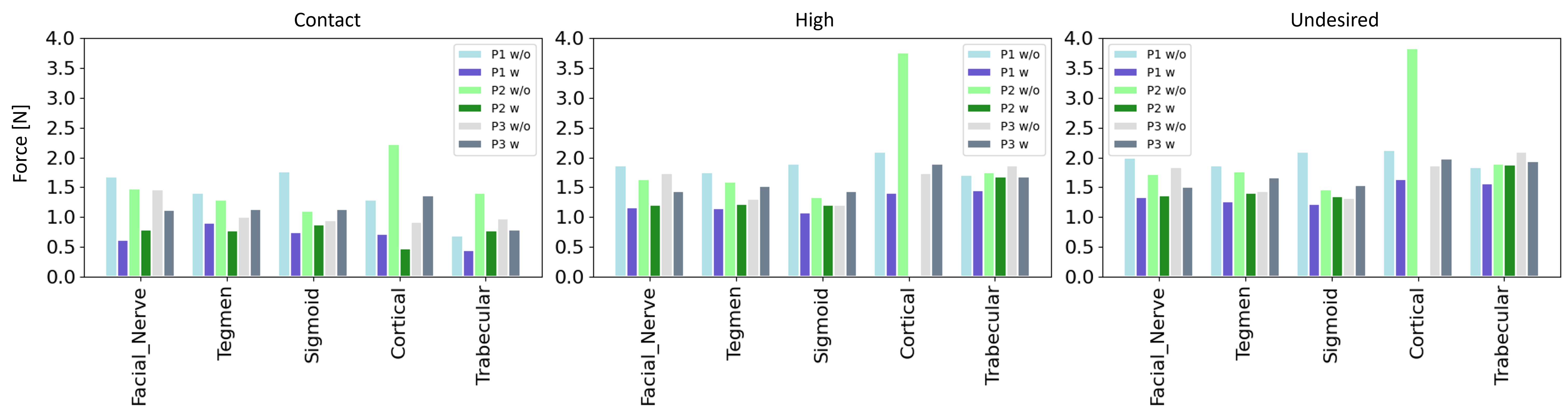}
\caption{\rev
{Comparison of the interaction force with and without proposed method. Contact, High and Undesired denote when the tissue interaction force being ($F_T > \rev{C} = 0.3 $[N]), ($F_T > \lambda_n $[N]) and ($F_T > \lambda_n +0.2 $[N]) respectively. Lower value indicates better result.}}\label{fig:Result}
\end{figure}

\rev{Throughout the experiment, no chattering or jittering was observed during the drilling (Fig. \ref{fig:JitterResult}).} During assessment of the experiment, we focused on four key metrics: statistics related to undesired forces($F_T > \lambda_n +0.2 $[N]
), statistics related to high forces($F_T > \lambda_n $[N]), statistics of the average interaction force ($F_T > \rev{C} = 0.3 $[N]), and the duration of time spent above the safety limit. The results of the proposed control method are listed in \rev{Fig. \ref{fig:Result}.}
The results show that the proposed method can effectively limit the interaction forces acting above the predefined safety limits \rev{for P1 and P2. Furthermore, it significantly reduced the proportion of time spent above the established safety limit, considering the total duration of drilling.}
\rev{However, for P3, while operating around the Tegmen, Sigmoid, and Corticol, there was an increase in interaction forces. This increase can be attributed to poor hand-eye coordination while performing the surgical task of skeletonizing the structure, as P3 was observed to make sudden, fast interactions with the anatomy during the experiment. Nonetheless, these findings highlight a limitation of our proposed method when users make sudden unintended interactions with the anatomy, emphasizing the importance of users possessing basic knowledge of the surgical task.
It is worth noting that these sudden, undesired interactions can be regulated by employing virtual fixtures around critical anatomical structures \cite{ishida2024haptic}, suggesting a potential avenue for addressing such challenges in future implementations.}

\begin{table}[t]
\caption{Tabular results for the average proportion of time spent above the safety limit.}\label{table:duration}%
\begin{tabular}{p{0.05\textwidth}p{0.125\textwidth}p{0.125\textwidth}p{0.125\textwidth}p{0.125\textwidth}p{0.125\textwidth}}
\toprule
Assist & Facial Nerve  & Tegmen & Sigmoid & Cortical & Trabecular \\
\midrule
w/o & 0.726 &  0.549 & 0.567 & 0.372 & 0.209 \\
w  & \textbf{0.322}  & \textbf{0.370} & \textbf{0.382} & \textbf{0.243} & \textbf{0.042} \\

\botrule
\end{tabular}
\end{table}

\section{Conclusion}\label{sec:conclusion}

In this study, we have introduced an adaptive force control method designed to actively limit interaction forces within the challenging surgical environment of skull base procedures. Our proposed situational-aware force control empowers the robot to execute constrained motions in the presence of elevated interaction forces while allowing the user to operate freely when forces are within safe operating ranges. \rev{Additionally, it enhances the sensation of touch when making contact with anatomical structures.} This adaptability is achieved through the automatic adjustment of admittance gains based on real-time assessments of contextual information within the DT environment.
The results from our initial experiments, involving a medical student \rev{and two engineering students} working with a cadaveric temporal bone, clearly demonstrate the effectiveness of our proposed method in successfully reducing undesired tool-tissue interactions.

Our future research will encompass several key areas:
(1) Threshold Refinement: While our initial study provided threshold values, we recognize the need for a more extensive investigation involving a larger sample to refine these thresholds. This will ensure that our system's safety limits are well-calibrated to meet the diverse demands of clinical practice;
(2) Integration of Safety-Driven Virtual Fixtures: We will explore the integration of safety-driven virtual fixtures aimed at aiding users in avoiding critical structures. This additional layer of safety guidance will further enhance the precision and safety of robotic-assisted surgical procedures;
(3) Comprehensive User Studies: We plan to rigorously evaluate the system's performance through extensive user studies, involving participants with varying levels of surgical experience. We aim to comprehensively assess how this technique contributes to the overall surgical experience, particularly in terms of safety and precision.





\backmatter

\bmhead{Supplementary information}

A supplementary video is provided with the submission. 

\bmhead{Acknowledgments}
Nimesh Nagururu is supported in part by NCATS TL1 Grant TR003100.
This work was also supported in part by a research contract from Galen Robotics, by NIDCD K08 Grant DC019708, by a research agreement with the Hong Kong Multi-Scale Medical Robotics Centre, and by Johns Hopkins University internal funds.

\section*{Declarations}

Under a license agreement between Galen Robotics, Inc and the Johns Hopkins University, Russell H. Taylor and Johns Hopkins University are entitled to royalty distributions on technology that may possibly be related to that discussed in this publication. Dr. Taylor also is a paid consultant to and owns equity in Galen Robotics, Inc. This arrangement has been reviewed and approved by Johns Hopkins University in accordance with its conflict-of-interest policies.
\bibliography{sn-bibliography}

\end{document}